\documentclass[runningheads]{llncs}
\usepackage[T1]{fontenc}

\usepackage{amsmath}
\usepackage{amsfonts}
\usepackage[utf8]{inputenc}
\usepackage{hyperref}
\usepackage{cleveref}
\usepackage{enumitem}
\usepackage{comment}
\usepackage{circledsteps}
\usepackage{subcaption}
\usepackage{multirow}
\usepackage{booktabs} 
\usepackage{graphicx}

\usepackage{color}

\usepackage{xspace}
\newcommand{\numruns}{10\xspace}

\begin{document}
\title{Evolutionary Reinforcement Learning for Interpretable Decision-Making in Supply Chain Management}
\titlerunning{Evolutionary Interpretable Reinforcement Learning for the Supply Chain}
%
\author{
Stefano Genetti\inst{1}\orcidID{0009-0004-2417-0319} 
\and
Alberto Longobardi\inst{2}
\and
Giovanni Iacca\inst{1}\orcidID{0000-0001-9723-1830
}}
\authorrunning{S. Genetti et al.}
%
\institute{
University of Trento\\
Via Sommarive 9, 38123 Povo (Trento), Italy\\
\email{\{stefano.genetti,giovanni.iacca\}@unitn.it}
\and
Adige BLM Group\\
Via per Barco 11, 38056 Levico Terme (Trento), Italy
\email{alberto.longobardi@blmgroup.it}
}
\maketitle              
\begin{abstract}
In the context of Industry 4.0, Supply Chain Management (SCM) faces challenges in adopting advanced optimization techniques due to the ``black-box'' nature of most AI-based solutions, which causes reluctance among company stakeholders. To overcome this issue, in this work, we employ an Interpretable Artificial Intelligence (IAI) approach that combines evolutionary computation with Reinforcement Learning (RL) to generate interpretable decision-making policies in the form of decision trees. This IAI solution is embedded within a simulation-based optimization framework specifically designed to handle the inherent uncertainties and stochastic behaviors of modern supply chains. To our knowledge, this marks the first attempt to combine IAI with simulation-based optimization for decision-making in SCM. The methodology is tested on two supply chain optimization problems, one fictional and one from the real world, and its performance is compared against widely used optimization and RL algorithms. The results reveal that the interpretable approach delivers competitive, and sometimes better, performance, challenging the prevailing notion that there must be a trade-off between interpretability and optimization efficiency. Additionally, the developed framework demonstrates strong potential for industrial applications, offering seamless integration with various Python-based algorithms.
\end{abstract}


\section{Introduction}
\label{sec:intro}

Disruptive technological innovations have consistently driven industrial revolutions, transforming production paradigms to enhance productivity~\cite{yin2018evolution}. In line with this trend, recent technological enhancements such as the Internet of Things
, big data analytics
, cloud computing
, robotics
, artificial intelligence
, and augmented reality
are motivating and enabling a \textit{fourth industrial revolution}
, commonly referred to as \textit{Industry 4.0}~\cite{lasi2014industry}. Since its introduction, the subject has received increasing attention from both academics and companies
which are increasingly compelled to adopt new technologies to maintain their competitiveness~\cite{adamson2017cloud}. Notably, competition is increasingly supply chain-based rather than company-specific~\cite{lambert2017issues}, 
with supply chains defined as interconnected networks that produce and deliver value~\cite{
mentzer2001defining}. In this domain, \textit{Supply Chain Management} (SCM) has emerged as a key element of Industry 4.0~\cite{brettel2014virtualization}. SCM involves the strategic coordination of business functions across companies to enhance both individual and collective performance~\cite{mentzer2001defining}.
Traditional decision-making approaches in SCM, often based on intuition and limited data, struggle in today’s dynamic environment.
On the other hand, modern decision-making, supported by \textit{business intelligence}, leverages data and analytics for better performance and competitive advantage.
While AI  techniques have proven effective for this task, they often lack transparency, making their decision processes hard to interpret~\cite{carter2023review,monteiro2023multi}. This misaligns with the interests of company stakeholders, who are hesitant to trust and deploy black-box AI-based solutions~\cite{dovsilovic2018explainable}. Instead, they tend to prefer solutions that reveal causal relationships and are perceived as more trustworthy and maintainable~\cite{arrieta2020explainable}. To address this issue, \textit{interpretable} and \textit{explainable} AI (IAI/XAI) AI solutions may provide a better alternative.

In this work, we adapt the IAI methodology from~\cite{custode2023evolutionary}, in the following referred to as \textsc{eldt} (Evolutionary Learning Decision Trees), for supply chain optimization. We integrate this approach within a simulation-based optimization framework which allows us to model complex, uncertain supply chain scenarios. We compare the performance of \textsc{eldt} against alternative approaches based on heuristics, metaheuristics, and Reinforcement Learning (RL) across two case studies, including a real-world application in laser-cutting machine production. The results show that \textsc{eldt} consistently provides solutions comparable to, and in some cases surpass, those of its competitors.

The rest of the paper is structured as follows. The next section reviews the background principles and the main related works. \Cref{sec:methods} describes the proposed approach. \Cref{sec:problems} defines the two supply chain optimization problems used to assess the approach. \Cref{sec:experimentalSetup} details the experimental setup. \Cref{sec:results} presents the experimental results. Finally, \Cref{sec:conclusions} concludes this work.


\section{Background and related work}
\label{sec:background}

Recent advancements, including affordable computational power, have led to the adoption of AI in industrial processes
, including SCM. 
Solving complex optimization problems in real-world organizations 
using exact methods is often infeasible, leading to the emergence of heuristics
, metaheuristics
, and deep learning approaches as a viable alternative~\cite{paternina2008simulation}. 
While these models can effectively support decision-making, they are often ``black-box''~\cite{hong2020human}, meaning their internal workings are not visible or comprehensible, which makes it difficult to trace how decisions are made. This lack of transparency hinders their broader adoption~\cite{carter2023review}, 
as managers and stakeholders remain accountable for outcomes produced by these models~\cite{monteiro2023multi}. For instance, a deep neural network might perform well during training, but it can struggle when deployed, necessitating extensive testing for trustworthiness~\cite{carter2023review}. Still, if an error occurs, it would be highly impractical to identify the root cause within the model, potentially resulting in substantial financial losses and safety hazards~\cite{baryannis2019predicting,heydarbakian2022interpretable,zaoui2023viability}. Moreover, the lack of transparency becomes a critical issue when the algorithms must be audited or when regulatory authorities, such as the European Union, mandate that outcomes must be understandable.
Ultimately, understanding causality is essential for decision-makers, enabling them to grasp cause-effect relationships and make informed choices.

The literature identifies two primary solutions to address this limitation: \textit{Explainable AI} (XAI) and \textit{Interpretable AI} (IAI)~\cite{arrieta2020explainable}.
XAI refers to techniques and methods that make the behavior of AI systems more understandable, providing insights into how a model reaches its decisions and thereby enhancing trust and transparency. 
IAI focuses on creating inherently understandable models, such as decision trees (DTs), which allow for direct inspection and assessment of security and safety. Furthermore, interpretable models enable users to uncover new correlations, identify relevant features, and leverage domain expertise to optimize model performance.
Despite their relevance for industrial decision-making, there is a noticeable lack of studies in the literature that propose XAI or IAI techniques for industrial optimization~\cite{ahmed2022artificial}, particularly in SCM~\cite{kosasih2024review}. 
One contributing factor is the widely accepted notion that there is a trade-off between interpretability and performance~\cite{baryannis2019predicting,bertolini2021machine}, though this trade-off remains unproven~\cite{rudin2019stop}. 
Additionally, the relatively novel adoption of AI in supply chains has led researchers to prioritize exploring its potential performance improvements rather than focusing on explaining its outputs and decisions~\cite{kosasih2024review}.

One of the few directions that have been explored concerning the use of IAI for SCM regards demand forecasting: for instance, in ~\cite{park2018development} the authors employ a \textit{neuro-symbolic} AI approach
which combines neural networks with symbolic reasoning, enhancing interpretability by expressing neural decisions through logical rules~\cite{kosasih2024review}.
In~\cite{lin2017incorporated}, the authors focus instead on supply chain risk management and utilize Ant Colony Optimization (ACO) to tackle the opacity of neural networks. In this case, the algorithm extracts intuitive rules (e.g., if-then-else statements) effectively representing the knowledge embedded in the black-box model.
One notable disadvantage of these implementations is that they introduce explainability by approximating or simulating the behavior of a given black-box model being used, rather than representing the model \textit{exactly}, which is instead the primary feature of IAI models. In~\cite{heydarbakian2022interpretable}, the authors introduce a framework using DTs, which are inherently interpretable, to classify supply chain suppliers based on their resilience capacity. Similarly,~\cite{baryannis2019predicting} emphasizes the synergy between AI models and supply chain experts in risk prediction. They compare a non-interpretable support vector machine (SVM) with an interpretable DT-based model for predicting delivery delays. The results show that while DTs achieve acceptable results, their displayed performance is inferior with respect to SVMs. The authors conclude that if precision is critical (e.g., where late deliveries incur significant costs) the weaker performance of the interpretable model may be unacceptable. Conversely, if understanding the factors behind delays is prioritized, a slight reduction in predictive accuracy might be acceptable.


\section{Proposed approach}
\label{sec:methods}

\begin{figure}[ht!]
    \centering
    \includegraphics[width=\linewidth]{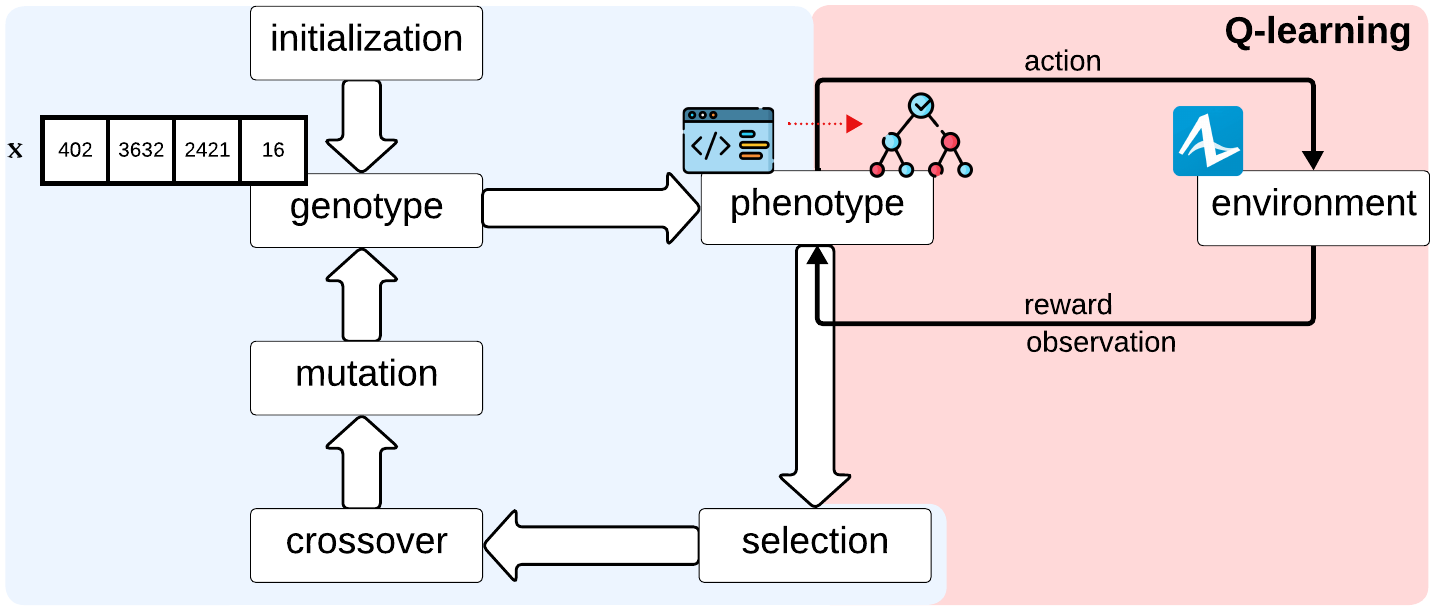}
    \caption{Conceptual scheme of the proposed algorithm's internal workings. Blue blocks indicate components related to the evolutionary loop, while red blocks represent elements of the \textit{Q}-learning loop. Adapted from~\cite{custode2023evolutionary}.}
    \label{fig:methodology}
\end{figure}

This work builds on the approach proposed in~\cite{custode2023evolutionary}, leveraging a two-level optimization scheme that combines evolutionary computation with RL to train interpretable policies in the form of DTs. 
In this method, the input to a DT, namely a one-dimensional feature vector, traverses the tree from the root through decision nodes until it reaches a leaf, which outputs the model's decision. Two key challenges arise in designing a DT for decision-making: \Circled{1} determining the conditions for splits at non-terminal nodes, and \Circled{2} assigning the appropriate action at each leaf. In the aforementioned two-level optimization scheme, the outer loop uses Grammatical Evolution~\cite{ryan1998grammatical} to evolve the DT structure, searching for optimal state space partitioning. The inner loop, instead, applies \textit{Q}-learning to map each aggregated state (i.e., all states that lead to a particular leaf) to an appropriate action.
\Cref{fig:methodology} shows a block diagram of the algorithm.

\noindent \textbf{Individual encoding.}
Each individual in the population is represented by a fixed-length list of integers, where each integer ranges from $0$ to $g_{\text{max}}$. The parameter $g_{\text{max}}$ is set much higher than the total number of productions in the Backus-Naur Form (BNF) grammar that governs the evolutionary process, ensuring uniform selection of all productions. Each genotype has a fixed length, and the initial population is generated randomly.\\
\noindent \textbf{Mutation operator.}
To promote population diversity, uniform mutation is applied to each individual, such that each gene has a probability $m_p$ of being replaced with a random value between $0$ and $g_\text{max}$.\\
\noindent \textbf{Crossover operator.}
To generate new offspring by combining the genotypes of individuals across generations, one-point crossover is used. In this method, a random point is chosen along the parent genotypes, splitting them into two parts. The offspring are then created by exchanging, after this point, the segments between the two parents. This crossover is applied to selected parent pairs in each generation, with uniform probability.\\
\noindent \textbf{Replacement of the individuals.}
Each generation, a new population replaces the current one using steady-state selection, ensuring that strong individuals are preserved throughout the evolutionary process.\\
\noindent \textbf{Fitness evaluation.}
The list of integers contained in each individual's genotype is translated into its corresponding phenotype, i.e., a DT-based policy, by applying the production rules of a problem-dependent grammar, following the procedure outlined in~\cite{custode2023evolutionary}. If the genotype lacks sufficient genes during this translation, the individual is penalized with a low fitness score.
Once the DT is created, its leaves are initialized with a set of actions, which are assigned a random value between $-1$ and $1$. This DT defines the policy $\pi$, which is used by an RL agent to interact with an environment implemented using OpenAI Gym\footnote{\url{https://github.com/openai/gym.git}}.

In each of the $e$ episodes, $\pi$ takes a state as input and selects an action at a leaf. To balance exploration and exploitation, action selection follows an $\epsilon$-greedy strategy: with probability $1-\epsilon$, the action with the highest value is chosen, otherwise a random action is selected. Feedback from the environment is used to update the corresponding leaf using Bellman's equation for \textit{Q}-learning:
$$Q(s,a) \gets (1-\alpha)Q(s,a)+\alpha(r+\gamma \underset{a'}{\text{max}}\{Q(s',a')\})$$
where $Q(s,a)$ is the value of action $a$ at state $s$, $\alpha$ is the learning rate, $r$ is the reward, $s$ and $a$ refer to the current state and action, and $s'$ and $a'$ refer to the next state and action, respectively.
The fitness assigned to the individual is the mean of the returns obtained across the $e$ episodes:
$$f(\pi) = \frac{1}{e}\sum_{i=1}^{e} R_i(\pi) = \frac{1}{e}\sum_{i=1}^{e} \sum_{k=1}^{T} r(s_k^i, \pi(s_k^i))$$
Here, $f(\pi)$ is the fitness of the agent encoding the policy $\pi$, $R_i(\pi)$ is the return obtained in the $i$-th episode (i.e., the sum of the rewards in the $i$-th episode), $T$ is the number of steps per episode, $r$ is the reward function, $s_k^i$ is the $k$-th state in the $i$-th episode, and $\pi(s_k^i)$ is the action taken by the policy in the state $s_k^i$.

\subsection{Simulation based optimization}
This work aims to apply optimization/RL methods for SCM. Due to the inherent complexities of supply chains such as uncertainties, stochastic behaviors, and high dimensionality, creating an analytical solution for diverse industrial applications is impractical. To address this, we propose a flexible simulation-optimization framework consisting of two key components: a simulation module, and an optimization module. 
The main design goals are applicability across various industrial scenarios, seamless integration with optimization/RL algorithms, and reasonable runtime performance.


For the simulation module, we adopt AnyLogic\footnote{\url{https://www.anylogic.com/}}, a Java-based simulation software widely used to develop high-fidelity system simulations across various domains, including supply chains
, business processes
, transportation
, and healthcare
. Its established role in recent research further supports our choice~\cite{ivanov2016operations,afanasyev2023system,wartelle2023study}. The optimization module leverages algorithms written in Python, chosen for its broad adoption and availability of state-of-the-art implementations. 

Communication between the AnyLogic's Java simulation environment and the Python-based optimizer is enabled by the open-source package ALPypeOpt\footnote{\url{https://github.com/MarcEscandell/ALPypeOpt.git}}.
Specifically, communication is achieved by implementing two methods from the \texttt{ALPypeOptClientController} interface: \texttt{setupAndRun}, which executes simulations using the decision variables provided by the optimizer, and \texttt{getModelOutput}, which allows the retrieval of the simulation results, such as revenue or production costs. Through an \texttt{AnyLogicModel} object, the optimizer interacts with AnyLogic, specifying inputs, triggering simulations, and retrieving outputs. 
In the optimization loop, the optimizer explores the solution space by proposing decision variable assignments, sends these to AnyLogic, and uses the simulation output to evaluate the objective function and guide the search. 


\section{Use cases}
\label{sec:problems}

We evaluated the proposed simulation-based optimization framework outlined in \Cref{sec:methods} in two supply chain problems, namely a make-or-buy decision problem and a hybrid flow shop scheduling problem, which are detailed below.

\noindent \textbf{Make-or-buy decision problem.} The make-or-buy decision, also known as the \textit{outsourcing decision}, is a common and critical task in SCM and industrial optimization~\cite{mcivor2000case}. 
In recent years, there has been growing recognition of the benefits associated with shifting from traditional in-house production to an outsourcing approach, where certain operations previously performed internally are delegated to external suppliers~\cite{canez2000developing}. The ability to accurately determine which activities to retain in-house and which to outsource can significantly impact a company's profitability and competitive standing~\cite{paul1994make}. Outsourcing certain production stages instead of handling them internally can lead to reduced production times, savings in physical space, and simplification of business processes.
However, the decision of what to make and what to buy is complex, as evidenced by the extensive literature proposing various frameworks to guide this process~\cite{serrano2018should}. While outsourcing can offer flexibility and help companies respond to rapid market changes,
it also carries risks, such as unexpected cost increases and heightened dependency on a broad range of suppliers~\cite{mcivor1997strategic}. As pointed out in~\cite{blaxill1991fallacy}, many companies lack a solid foundation for making these decisions, often relying on overhead costs as the primary factor guiding their strategy. A more holistic view that considers additional contributing aspects is necessary for making sustainable, long-term business decisions. In view of this, the use of interpretable AI can provide a significant contribution.

In this study, we examine an instance of the make-or-buy decision problem which can be further adapted to address more specific scenarios. Our focus is on the supply chain of a fictional company aiming to fulfill customer orders efficiently. Each order requires the assembly of a specified number of components A, B, and C, and has a given deadline. In the supply chain, four plants are involved: Plant A produces component A, Plant B produces component B, Plant C produces component C, and Plant D assembles the components to complete orders. The company can fulfill orders using its own plants or outsource the production. Outsourcing ensures timely completion but comes with a significant cost of $30$ (generic cost units) per order. The primary objective of this single-objective optimization problem is to maximize the total revenue $R$, which is defined as:
${R = 100 \times n_{ordersOnTime} + 50 \times n_{ordersOutOfTime} - 30 \times n_{ordersOutsourced}}$. 

On-time deliveries earn the highest reward ($100$) for customer satisfaction. Late deliveries still contribute ($50$), acknowledging the value of completion despite delays. Outsourcing is penalized ($-30$) due to added costs and reduced control. This set of coefficients prioritizes efficient order completion, encouraging timely, internal fulfillment while minimizing delays and outsourcing.
The AnyLogic simulation models uncertainties and stochastic behaviors, reflecting real-world variability in production and logistics that cannot be easily addressed through a purely analytical approach. These uncertainties are modeled by allowing production times at Plants A, B, and C to be not fixed but follow probability distributions. A truck cyclically visits these companies to transport components to Plant D (assembly unit). If no components are ready for pickup, the truck skips the stop. Loading, unloading, and travel times are also stochastic, with values sampled from uniform distributions within given ranges.

Given a number of orders $n_{orders}$, the decision variables are a set of binary values ${\mathbf{x}=[x_1, x_2, \hdots, x_{n_{orders}}]}$, where $x_i=1$ indicates outsourcing an order, and $x_i=0$ indicates fulfilling it using the internal supply chain.
The objective is to find the optimal set of binary decisions $\mathbf{x}^*$ that maximizes the total revenue $R=f(\mathbf{x})$ obtained from the AnyLogic simulation given the outsourcing decisions.
Even for small problem instances, the number of possible realizations of $\mathbf{x}$ is considerably large. For example, with an input of $100$ orders, there are $2^{100}$ possible arrangements. This renders brute-force methods computationally infeasible, further justifying the need for AI-based solutions to explore this vast solution space and identify regions that yield good revenues.

\noindent \textbf{Hybrid flow shop scheduling problem.} Scheduling problems are ubiquitous across several application domains like manufacturing, services, and project management, each with its unique constraints and goals. In its simplest form, scheduling deals with assigning jobs to machines in a specific order to optimize a desired objective function. In more depth, the literature refers to \textit{flow shop scheduling} as the optimization problem in which multiple jobs need to be processed on a series of machines, where each job must pass through all machines in a predetermined order. \textit{Hybrid flow shop} (HFS) scheduling introduces an additional layer of complexity by allowing multiple machines at each stage of the production process. Unlike in a traditional flow shop, where each stage is served by a single machine, in HFS each stage can have multiple parallel machines. This reflects real-world scenarios where a production stage might have several identical or different machines working in parallel in order to introduce flexibility, increase capacities, and avoid bottlenecks in some operations that are more time-consuming~\cite{yu2018genetic}. As a result, HFS is a widely used production system that is applicable across various industrial sectors. For example, in automotive manufacturing, different parts of a vehicle might be assembled in parallel before being brought together in the final assembly line.

In its classical formulation, an HFS scheduling problem~\cite{ruiz2010hybrid} involves a set of $n$ input jobs ($j_1,j_2,\hdots,j_n$), that must be processed sequentially through a workshop consisting of $m$ stages. Each stage $i$ may contain $m_i \geq 1$ machines operating in parallel and for at least one stage $m_i > 1$. The production flow is the same for all jobs: they proceed in order through stage 1, stage 2, and so on until stage~$m$. All jobs and machines are available for processing from time zero, ensuring that no delays are caused by initial availability constraints. Within each stage, machines are identical in function but may have different processing times for different jobs. Each machine can process only one job at a time, and preemption is not allowed, meaning that once a job starts processing on a machine, it must continue until completion without interruptions. After finishing at one stage, a job is immediately available for processing at the next stage. Although jobs may wait between stages, the buffer space is assumed to be unlimited, so jobs can remain in the buffer until a machine becomes available. Given these constraints, the problem is to find a schedule that optimizes a given objective function. The most commonly used objective functions~\cite{li2011new} for the HFS problem, as identified in the literature, include the makespan
, the total flow time
, the maximum tardiness
, the total completion time
, the total tardiness. 
Makespan refers to the total time required to complete all jobs, i.e., the completion time of the last job. Total flow time is the sum of the time each job spends in the system, from its release to its completion. Total completion time represents the sum of the completion times of all jobs. Lastly, total tardiness measures the total sum of delays for all jobs beyond their respective due dates.
According to the results presented in~\cite{gupta1988two}, the HFS problem is NP-hard. This implies that it is infeasible to develop an algorithm that solves the problem with polynomial time complexity, even in the simplified scenario where there are two stages, one with two machines and the other with a single machine.

In this study, we consider a real-world application of the HFS problem. Our focus is on a supply chain scenario involving an Italian manufacturing company, a global leader in the production of laser-cutting machines for tubes. The objective of the supply chain is to produce and deliver laser-cutting machines in response to customer demand. The organization manufactures laser-cutting machines, categorized into two main product families with similar production cycles within each group. The ``LT7 family'' includes 4 machines, identified as LT7, LT7p, LT7 INS., and LT7p INS., while the ``LT8 family'' comprises 8 machines, identified as LT8, LT8p, LT8 ULA, LT8p ULA, LT8 12, LT8p 12, LT8 12 ULA, and LT8p 12 ULA. Each machine type has unique production stages with varying durations and specific human resource requirements, classified as M, E, and R. The company has multiple human resources for each stage, so it is possible to work on more than one laser-cutting machine at the same time. The manufacturing facility has 20 assembly areas, allowing for simultaneous work on up to 20 machines. Completed machines are transported by truck to the logistics centers, with transportation time being a significant factor.

Within this scenario, it is crucial to assign proper priorities to customer laser-cutting machine orders to efficiently allocate human resources to machines on a daily basis, thereby optimizing the overall supply chain. From this perspective, we can view this situation as an instance of the HFS problem applied to a real-world supply chain scenario. An instance of the HFS problem in this context involves a sequence (potentially with more than 20 entries) of laser-cutting machine orders (the jobs to be processed). Each order is characterized by the following attributes: \Circled{1}~order ID; \Circled{2}~machine type (\texttt{mt}); \Circled{3}~due date (\texttt{dd}); \Circled{4}~basement arrival date (\texttt{db}); \Circled{5}~date of electrical panel arrival (\texttt{de}). The basement arrival date marks when the first M phase can be processed. In other words, it is not possible to start the sequence of tasks with people belonging to stage M before this date. Similarly, the electrical panel arrival date indicates when the first E phase can begin, preventing E tasks from starting before this date. Additionally, the problem instance specifies the total number of available human resources for each type (M, E, R). The objective of the optimizer is to determine the optimal job schedule.
We adopt an indirect encoding scheme that represents solutions as job permutations, omitting some decision variables required for constructing an HFS schedule. Paired with surrogate heuristics, this approach reduces the search space and yields better results than direct encoding~\cite{urlings2010genetic,yu2018genetic}. Job permutations are decoded into schedules using a \textit{list scheduling} technique~\cite{yu2018genetic}.
Among the common metrics used in the literature to evaluate the output of an HFS solver, we focus on minimizing the makespan, which intuitively reflects the efficiency of other metrics as well. 


\section{Experimental setup}
\label{sec:experimentalSetup}

\noindent \textbf{Computational setup.} For our experiments, we used the free Personal Edition of AnyLogic, which does not support exporting a simulation model as a standalone application. As a result, the simulations had to be executed directly through the AnyLogic GUI. The experiments were conducted on a Windows 11 laptop with a 14-core Intel i7-12700H CPU @ 2.30 GHz and 32GB of RAM. The total execution time for all experiments was ca. 400 CPU core hours. Our AnyLogic simulation model and Python code are publicly available on GitHub.\footnote{\url{https://github.com/DIOL-UniTN/eldt-scm}}.

\noindent \textbf{Algorithm settings.} The optimization algorithms we implemented are categorized into two groups, namely:
\begin{itemize}[leftmargin=*]
    \item \textit{Schedule-as-a-whole approaches}. These methods operate on the entire list of orders as a whole. We implemented a Random Search (\textsc{rs}) baseline and included \texttt{optuna}\footnote{\url{https://github.com/optuna/optuna.git}} (\textsc{optuna}), a popular optimization framework, configured with its default sampler (Tree-structured Parzen Estimator~\cite{watanabe2023tree}) 
    and pruner (Median Pruner). Additionally, we incorporated two metaheuristics, namely a Genetic Algorithm (\textsc{ga}) and Ant Colony Optimization (\textsc{aco}), both implemented using the \texttt{inspyred} Python library\footnote{\url{https://github.com/aarongarrett/inspyred.git}}. For the HFS problem, we also considered a greedy heuristic, \textsc{greedy}, which prioritizes jobs by the earliest due dates.
    \item \textit{Policy-generating approaches}. These methods focus on training a policy that processes one order at a time and makes decisions based on its features. In this category, we included a deep RL method (\textsc{rl}) based on Proximal Policy Optimization (PPO)~\cite{schulman2017proximal} using the \texttt{rllib}\footnote{\url{https://docs.ray.io/en/latest/rllib/index.html}} implementation. We also incorporated an IAI solution that evolves DTs using Genetic Programming (\textsc{gp}), implemented using the \texttt{deap} Python library\footnote{\url{https://github.com/DEAP/deap.git}}. Finally, our proposed approach (\textsc{eldt}), as described in \Cref{sec:methods}, belongs to this policy-generating category. 
\end{itemize}
For the implementation details, readers are referred to our GitHub codebase.

To ensure a fair comparison, each algorithm was allocated 5000 AnyLogic simulation executions. Hyperparameters affecting the computational budget were adjusted accordingly, while other parameters were set to their library defaults. For \textsc{eldt}, we used the settings from~\cite{custode2023evolutionary}. Due to space constraints, detailed hyperparameter descriptions are reported in the public GitHub repository.\\
\noindent \textbf{Datasets.} Because of the specificity of our scenarios, suitable benchmarks for evaluating the implemented algorithms are not readily available. As a result, for both problems, we generated datasets through stochastic procedures (which are also made available in the GitHub repository). For the HFS, we also included a real-world dataset with data provided by the company.

As regards the make-or-buy decision problem, we handcrafted a dataset consisting of 100 order entries. Each order requires a specified quantity of three components: type A, type B, and type C. The quantity of each component is sampled uniformly at random from the range $[0,20]$. Additionally, each order's deadline is generated stochastically by adding a random number of days, uniformly sampled between 800 and 1500, to the starting simulation date of June 19, 2024. Due to the stochastic nature of the generation process, to ensure a robust and reliable analysis, we generated three independent instances of this dataset, referred to as \texttt{list1}, \texttt{list2}, \texttt{list3}.

For the HFS scheduling problem, we generated five test instances using the procedures described below. In all experiments, the number of human resources of types M, E, and R available to work on laser-cutting machine orders is fixed at $M=E=R=5$, a value manually calibrated to avoid trivial solutions: too few or too many human resources would result in similar performance across all algorithms. \Circled{1}~Dataset \texttt{d1} includes a list of orders for all $12$ laser-cutting machines in both families. The arrival dates for both the basement component and the electrical panel are independently and uniformly distributed between day $1$ and day $20$. This relatively narrow range is calibrated by hand to ensure that the close arrival times of the components make scheduling decisions more impactful, thereby highlighting differences in algorithm performance. The delivery date for each order is set to the basement arrival date plus a random number between $20$ and $50$, which is consistent with the lead time of the company. This dataset contains $100$ laser-cutting machine orders. \Circled{2}~Dataset \texttt{d2} is generated using the same procedure as dataset \texttt{d1}, but it is limited to machines from the ``LT7 family''. \Circled{3}~Dataset \texttt{d3}, instead, only includes laser-cutting machines from the ``LT8 family''. \Circled{4}~Dataset \texttt{d4}, similarly to \texttt{d1}, includes orders for all $12$ laser-cutting machines in both families. However, differently from \texttt{d1}, \texttt{d4} simulates an annual production plan divided into five groups, $\text{G1}$ (days 1–73), $\text{G2}$ (days 74–146), $\text{G3}$ (days 147–219), $\text{G4}$ (days 220–292), and $\text{G5}$ (days 293–365). Each machine is randomly assigned to one of these groups. For each order in the dataset, the basement arrival date is randomly chosen within the corresponding group's day range, as is the electrical panel arrival date. The due date is set to the basement arrival date plus a random number between 20 and 50. The dataset consists of $100$ entries. \Circled{5}~Dataset \texttt{d5} is populated with $345$ orders directly provided by the company from 2021 to 2024, consisting of LT7 and LT8 machines.


\section{Results}
\label{sec:results}
To ensure reliable results, we aggregated data from \numruns independent runs, accounting for the algorithms' stochasticity. For the \textsc{greedy} algorithm, a single run is sufficient due to its deterministic nature.

\begin{figure}[ht!]
    \centering
    \begin{subfigure}{0.6\linewidth}
        \centering
        \includegraphics[width=\linewidth]{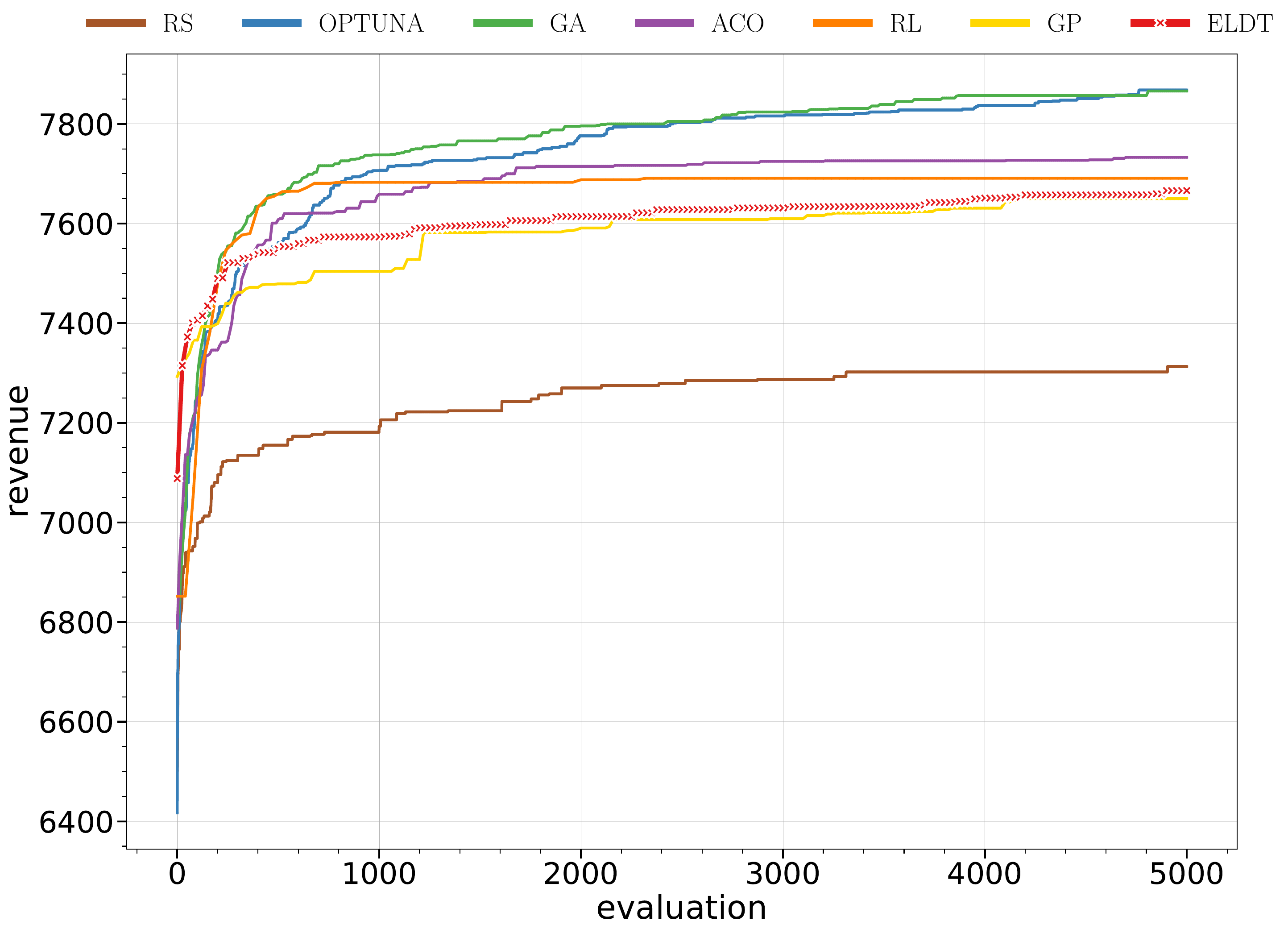}
        \caption{Best revenue $R$ (average across \numruns runs) found by each algorithm across evaluations.}
        \label{fig:fitnessTrentList2}
    \end{subfigure}
    \hfill
    \begin{subfigure}{0.35\linewidth}
        \centering
        \begin{subfigure}{\linewidth}
            \centering
            \includegraphics[width=\linewidth]{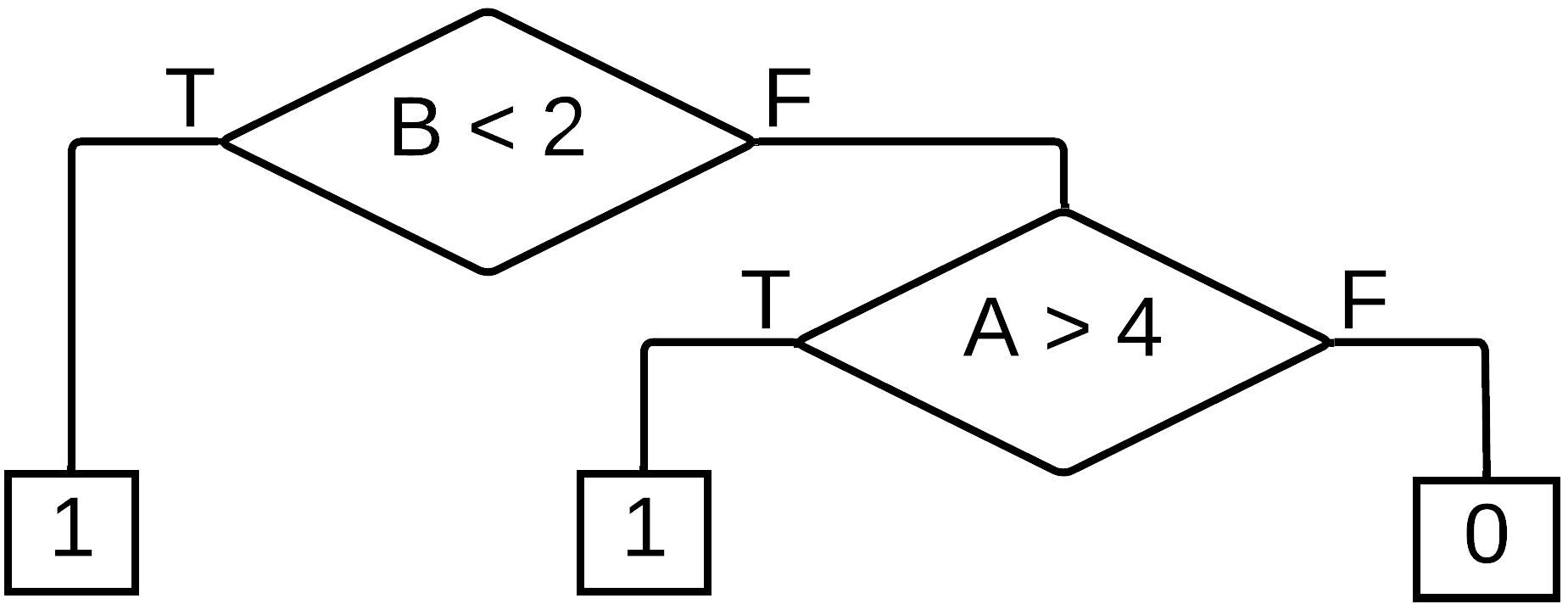}
            \caption{Best DT found by \textsc{eldt}.}
            \label{fig:outsourcingDtGeList2}
        \end{subfigure}
        \begin{subfigure}{\linewidth}
            \centering
            \includegraphics[width=\linewidth]{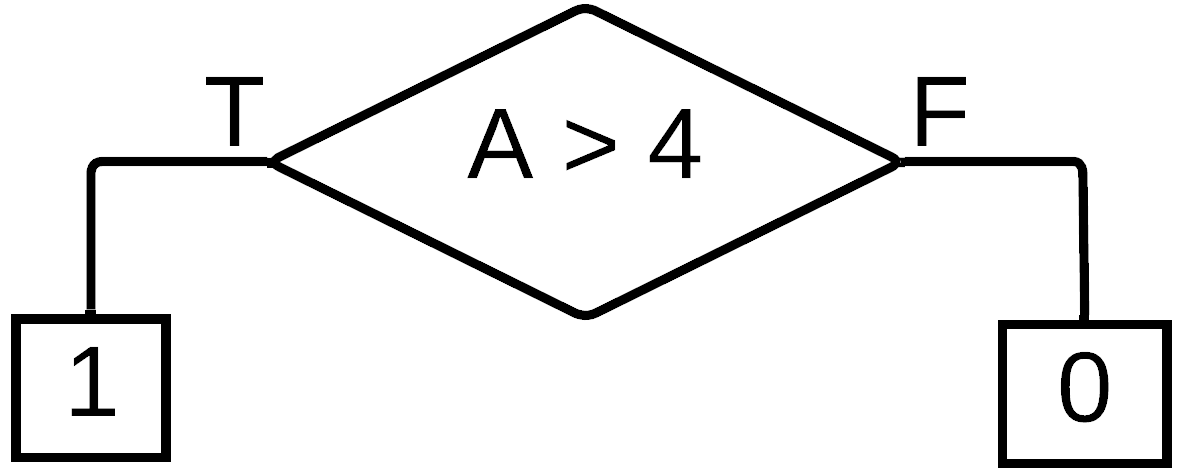}
            \caption{Best DT found by \textsc{gp}.}
            \label{fig:outsourcingDtGpList2}
        \end{subfigure}
    \end{subfigure}
    \caption{Results for the make-or-buy decision problem on \texttt{list1}.}
    \label{fig:outsourcingQUalitativeResults}
    \vspace{-0.3cm}
\end{figure}

Regarding the make-or-buy decision problem, \Cref{fig:outsourcingQUalitativeResults} shows the fitness trend achieved by the optimization algorithms on dataset \texttt{list1} (\Cref{fig:fitnessTrentList2}), along with the best DTs found by \textsc{gp} (\Cref{fig:outsourcingDtGpList2}) and \textsc{eldt} (\Cref{fig:outsourcingDtGeList2}). \Cref{tab:comparisonOutsourcing} provides a quantitative comparison of the results across the three datasets.

\begin{figure}[ht!]
    \centering
    \includegraphics[width=\linewidth]{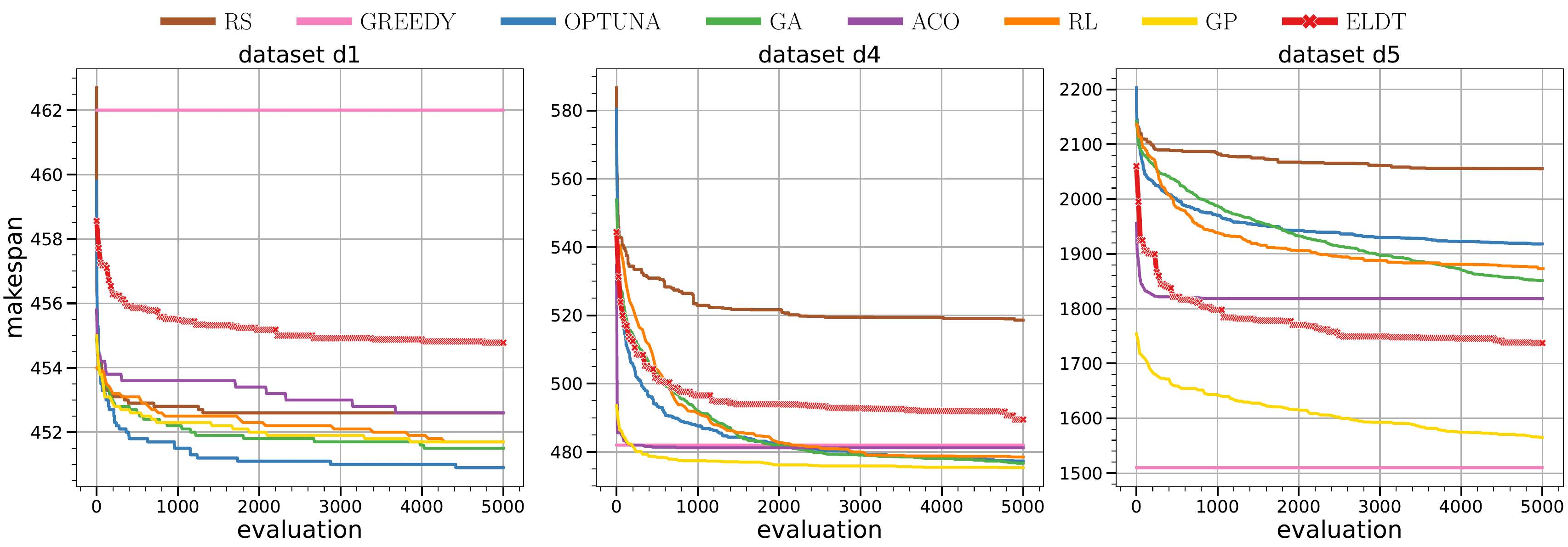}
    \caption{Results for the HFS scheduling problem (datasets \texttt{d1}, \texttt{d4}, and \texttt{d5}). The y-axis displays the best makespan (average across \numruns runs) found by each algorithm across evaluations. The result of the \textsc{greedy} policy is reported for reference.}
    \label{fig:hfsFitnessTrend}
    \vspace{-0.3cm}
\end{figure}

\begin{figure}[ht!]
    \centering
    \begin{subfigure}[t]{0.22\linewidth}
        \centering
        \includegraphics[width=\linewidth]{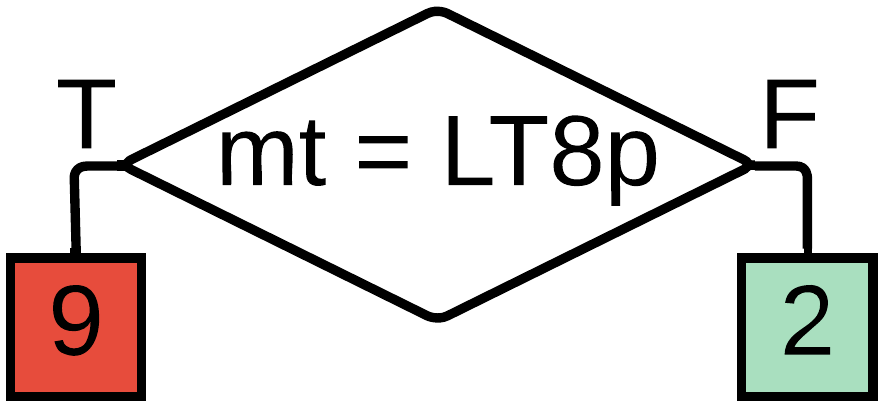}
        \caption{Best DT on \texttt{d1}.}
        \label{fig:hfsDtD1}
    \end{subfigure}
    \hfill
    \begin{subfigure}[t]{0.38\linewidth}
        \centering
        \includegraphics[width=\linewidth]{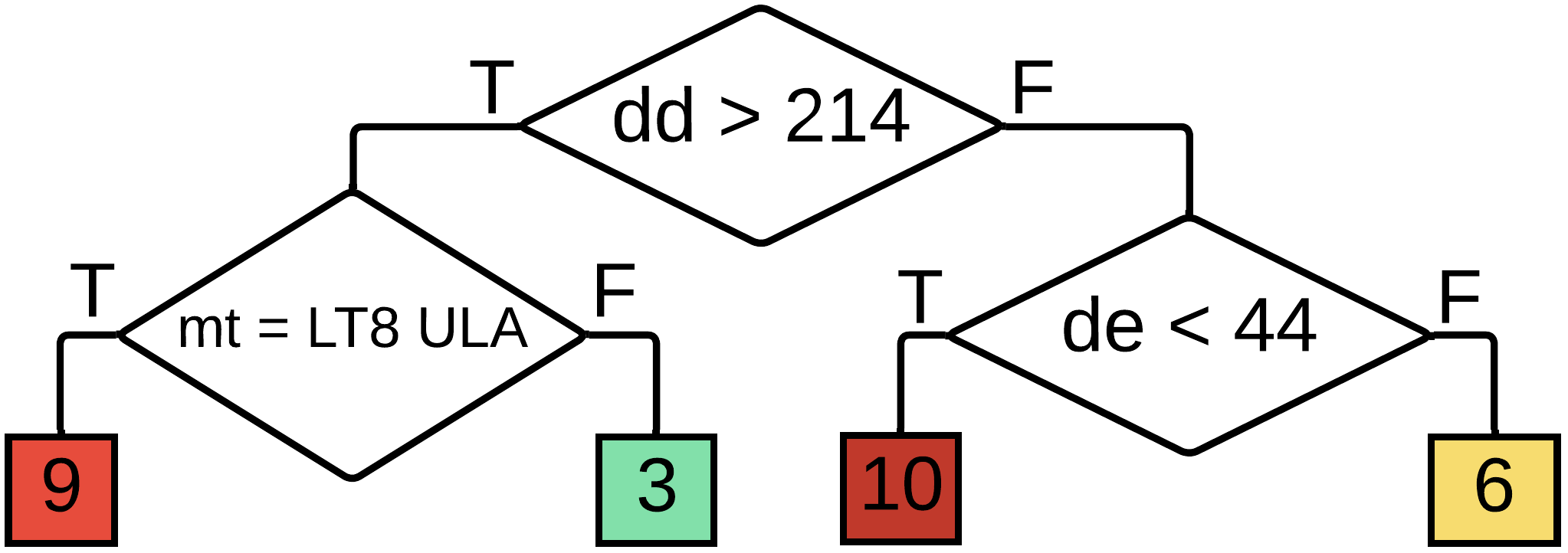}
        \caption{Best DT on \texttt{d4}.}
        \label{fig:hfsDtD4}
    \end{subfigure}
    \hfill
    \begin{subfigure}[t]{0.38\linewidth}
        \centering
        \includegraphics[width=\linewidth]{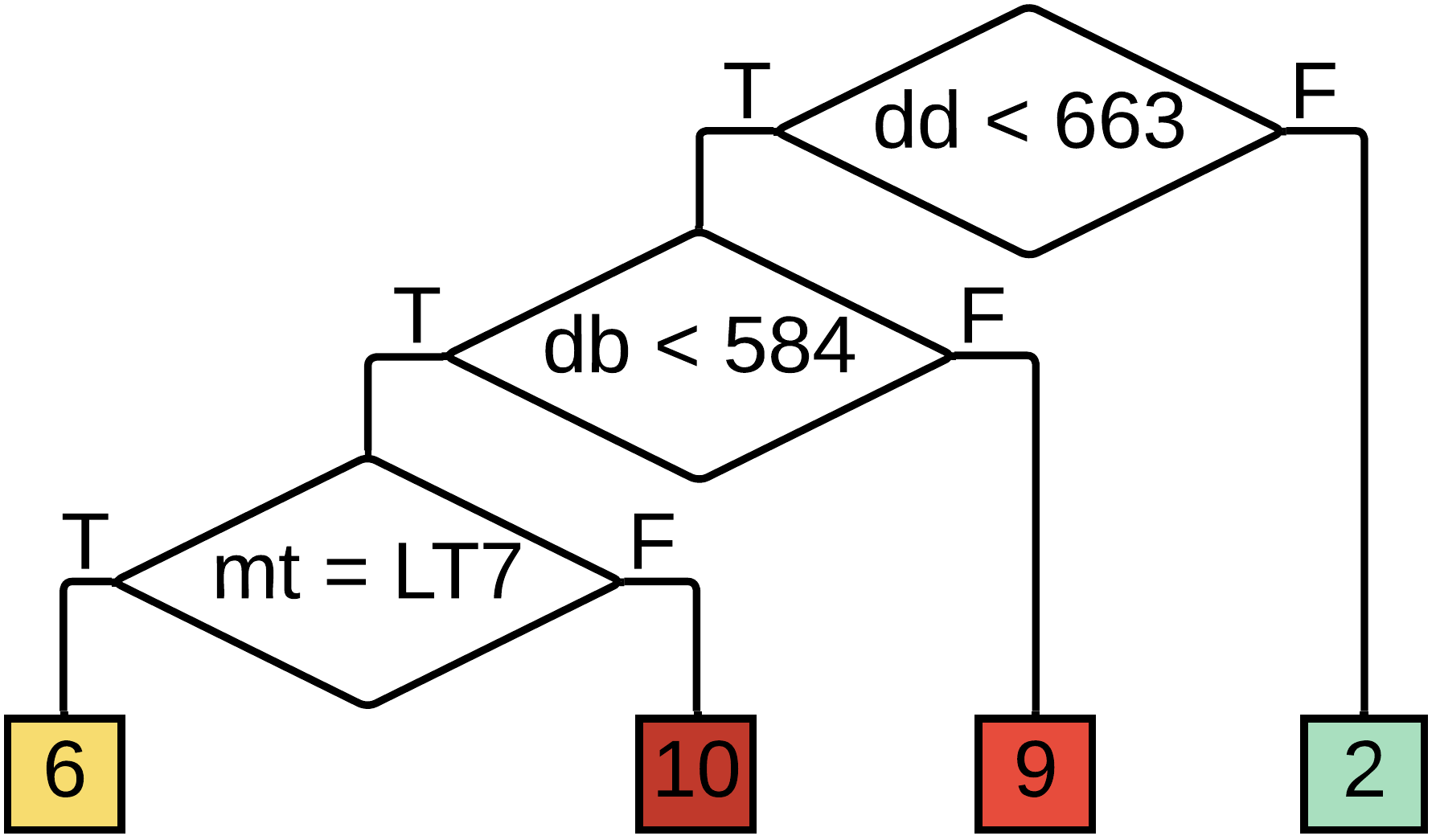}
        \caption{Best DT on \texttt{d5}.}
        \label{fig:hfsDtD5}
    \end{subfigure}
    \vspace{-0.2cm}
    \caption{Results for the HFS scheduling problem. Best DTs found by \textsc{eldt} on datasets \texttt{d1}, \texttt{d4}, and \texttt{d5} across \numruns runs, with leaf nodes colored on a gradient from light (low priority for input order) to dark red (high priority for input order), indicating how the simulation model prioritizes scheduling.}
    \label{fig:hfsDecisionTrees}
    \vspace{-0.5cm}
\end{figure}

For the HFS scheduling problem, \Cref{fig:hfsFitnessTrend} shows the fitness trends of the algorithms on datasets \texttt{d1}, \texttt{d4}, and \texttt{d5}. \Cref{fig:hfsDecisionTrees} presents the best DTs obtained by \textsc{eldt} for the same datasets.
\Cref{tab:comparisonHFS} provides the detailed results.

Note that, for illustration purposes, the DTs shown in the figures have undergone a post-processing step. This step is necessary when the trees contain subtrees that are never executed. While it would be possible to integrate a mechanism to prevent the generation of such non-executable branches within the algorithms, doing so would increase the computational complexity. Furthermore, retaining this redundancy in the population can promote greater diversity. 

The results demonstrate the effectiveness and flexibility of our simulation-based optimization framework. In particular, comparing the two categories of methods (schedule-as-a-whole and policy-generating), it emerges that neither consistently outperforms the other. However, one potential advantage of policy-generating approaches is that the policy obtained can be continuously refined based on feedback from the simulator as new orders arrive dynamically. In contrast, schedule-as-a-whole methods need to restart the optimization process each time the order list is updated.

\begin{table}[ht!]
\centering
\setlength{\tabcolsep}{5pt}
\vspace{-.6cm}
\caption{Results for the make-or-buy decision problem. Revenue $R$ (mean $\pm$ std. dev. across \numruns runs) of the different algorithms is reported. The best for each dataset is highlighted in bold. Asterisk indicates a statistically significant difference w.r.t. \textsc{eldt} results (Wilcoxon test, $\alpha=0.05$); $ns$ denotes no significance.}
\label{tab:comparisonOutsourcing}
\resizebox{\textwidth}{!}{
\begin{tabular}{cccccccc}
\toprule
\multirow{2}{*}{\textbf{Dataset}} & \multicolumn{4}{c}{\textbf{Schedule-as-a-whole approaches}} & \multicolumn{3}{c}{\textbf{Policy-generating approaches}} \\
\cmidrule(l{5pt}r{5pt}){2-5}
\cmidrule(l{5pt}r{5pt}){6-8}
& \textsc{random}  & \textsc{optuna} & \textsc{ga} & \textsc{aco} & \textsc{rl} & \textsc{gp} & \textsc{eldt} \\
\midrule
\texttt{list1} & ${7313.00 \pm 76.60}^*$ & ${\mathbf{7868.00 \pm 50.51}}^*$ & ${7866.00 \pm 25.91}^*$ & ${7733.00 \pm 37.73}^*$ & ${7691.00 \pm 26.44}^{\text{ns}}$ & ${7650.00 \pm 149.00}^{\text{ns}}$ & $7666.20 \pm 67.66$ \\
\texttt{list2} & ${7397.00 \pm 35.61}^*$ & ${7802.00 \pm 102.93}^*$ & ${\mathbf{7807.00 \pm 45.72}}^*$ & ${7704.00 \pm 34.38}^{\text{ns}}$ & ${7698.00 \pm 32.25}^{\text{ns}}$ & ${7734.00 \pm 62.40}^{\text{ns}}$ & $7712.60 \pm 61.03$ \\
\texttt{list3} & ${7286.00 \pm 58.73}^*$ & ${7848.00 \pm 65.29}^*$ & ${\mathbf{7858.00 \pm 28.60}}^*$ & ${7746.00 \pm 43.51}^*$ & ${7719.00 \pm 35.73}^{\text{ns}}$ & ${7752.00 \pm 102.83}^{\text{ns}}$ & $7662.80 \pm 79.71$ \\
\bottomrule
\end{tabular}
}
\end{table}

\begin{table}[ht!]
\centering
\setlength{\tabcolsep}{5pt}
\vspace{-1.2cm}
\caption{Results for the HFS scheduling problem. Average $\pm$ std. dev. of best makespan over \numruns runs across five datasets. Note that the \textsc{greedy} algorithm is deterministic. Bold highlights the best result per dataset. Asterisk indicates a statistically significant difference w.r.t. \textsc{eldt} (Wilcoxon test, $\alpha=0.05$).}
\label{tab:comparisonHFS}
\resizebox{\textwidth}{!}{
\begin{tabular}{ccccccccc}
\toprule
\multirow{2}{*}{\textbf{Dataset}} & \multicolumn{5}{c}{\textbf{Schedule-as-a-whole approaches}} & \multicolumn{3}{c}{\textbf{Policy-generating approaches}} \\
\cmidrule(l{5pt}r{5pt}){2-6}
\cmidrule(l{5pt}r{5pt}){7-9}
& \textsc{random} & \textsc{greedy} & \textsc{optuna} & \textsc{ga} & \textsc{aco} & \textsc{rl} & \textsc{gp} & \textsc{eldt} \\
\midrule
d1 & ${452.60 \pm 0.52}^*$ & ${462.00}^*$ & ${\mathbf{450.90 \pm 0.32}}^*$ & ${451.50 \pm 0.53}^*$ & ${452.60 \pm 0.52}^*$ & ${451.70 \pm 0.48}^*$ & ${451.70 \pm 0.48}^*$ & $454.78 \pm 0.74$ \\
d2 & ${436.90 \pm 0.32}^*$ & ${445.00}^*$ & ${\mathbf{436.00 \pm 0.00}}^*$ & ${436.30 \pm 0.48}^*$ & ${437.20 \pm 0.79}^*$ & ${436.80 \pm 0.42}^*$ & ${436.20 \pm 0.79}^*$ & $439.08 \pm 0.44$ \\
d3 & ${458.20 \pm 0.42}^*$ & ${461.00}^*$ & ${\mathbf{457.10 \pm 0.32}}^*$ & ${457.60 \pm 0.52}^*$ & ${458.00 \pm 0.67}^*$ & ${457.40 \pm 0.52}^*$ & ${457.70 \pm 0.48}^*$ & $460.88 \pm 0.53$ \\
d4 & ${518.60 \pm 4.48}^*$ & ${482.00}^*$ & ${477.30 \pm 1.77}^*$ & ${476.70 \pm 1.16}^*$ & ${481.20 \pm 2.62}^*$ & ${478.50 \pm 1.35}^*$ & ${\mathbf{475.40 \pm 0.70}}^*$ & $489.48 \pm 5.27$ \\
d5 & ${2055.40 \pm 10.34}^*$ & ${\mathbf{1510.00}}^*$ & ${1918.12 \pm 10.83}^*$ & ${1851.10 \pm 23.86}^*$ & ${1818.20 \pm 19.01}^*$ & ${1872.90 \pm 44.09}^*$ & ${1564.80 \pm 24.10}^*$ & $1737.30 \pm 36.65$ \\
\bottomrule
\end{tabular}
}
\vspace{-0.2cm}
\end{table}

It is worth noticing that, for the make-or-buy decision problem, the large standard deviation values in \Cref{tab:comparisonOutsourcing} reflect the high variability and uncertainty inherent in the simulated supply chain for the make-or-buy decision problem. On the other hand, for the case of HFS, see \Cref{tab:comparisonHFS}, it should be noted that for dataset \texttt{d2}, \textsc{optuna} likely achieved the optimal solution across all runs, as indicated by its zero standard deviation.

Notably, on both problems \textsc{eldt} achieved results comparable to (or in some cases better than) competitors within the given budget while producing simple, interpretable policies. In most cases, pairwise comparisons (Wilcoxon rank-sum test, $\alpha=0.05$) show that differences between \textsc{eldt} and the competitor solutions are statistically significant. In general, interpretable DTs approaches (\textsc{eldt} and \textsc{gp}) perform on par with, and sometimes outperform, black-box optimizers, particularly excelling on the real-world dataset (note that \textsc{greedy} is not black-box). On dataset \texttt{d5}, \textsc{greedy} outperforms other algorithms, likely because of its longer time frame (approximately three years) compared to the other datasets, which span only a single year. Over such an extended time frame, prioritizing orders strictly by their deadlines appears to be the most effective approach, especially under practical constraints like limited human resources. The analysis of the fitness curve in \Cref{fig:fitnessTrentList2} however suggests that, with more evaluations, other optimization algorithms may achieve performance levels comparable to \textsc{greedy}. Further investigation is needed to validate these observations.

Interestingly, the policies found by \textsc{eldt}  sometimes resemble those achieved by \textsc{gp}, suggesting convergence towards similar policies. For instance, looking at \Cref{fig:outsourcingDtGeList2} and \Cref{fig:outsourcingDtGpList2} it can be seen that both methods recommending outsourcing for $\text{A}>4$ and advising no outsourcing below this threshold. Future work should investigate the possible convergences between \textsc{gp} and \textsc{eldt} DTs.

The DTs from \textsc{eldt} for the HFS problem also provide useful insights. For dataset \texttt{d1}, the tree (\Cref{fig:hfsDtD1}) prioritizes orders for machines of type ``LT8p'', which have the shortest manufacturing time. For \texttt{d5} (\Cref{fig:hfsDtD5}), the root node emphasizes delivery dates, hence mirroring the \textsc{greedy} baseline which prioritizes orders by deadline. A future enhancement could involve comparing the policies found by \textsc{eldt} with the scheduling strategies adopted by company practitioners.

\section{Conclusions}
\label{sec:conclusions}
In this work, we applied optimization/RL algorithms to enhance decision-making in supply chain management, a key element of Industry 4.0. We addressed two supply chain optimization problems using a versatile simulation-based optimization framework that successfully integrated various Python-based algorithms. By adopting AnyLogic to model the supply chain, we simplified development by capturing real-world complexities like uncertainties and stochasticity. This approach enables non-experts in AI to contribute to model design and visualize optimization results, reducing the risks and costs of implementation.

We classified the tested methods into two groups: schedule-as-a-whole and policy-generating approaches. Among the latter, we considered, in particular, an instance of Genetic Programming along with the method proposed in~\cite{custode2023evolutionary}, which combines Grammatical Evolution with RL to generate interpretable DTs. 
Results show no clear winner between the two groups of methods, though policy-generating approaches have the potential advantage of continuously refining the decision-making process as new orders arrive dynamically, unlike schedule-as-a-whole approaches, which must reprocess the entire list for each new order.

To the best of our knowledge, this is the first attempt to apply a simulation-based optimization to generate interpretable models (based on DTs) for supply chain optimization. Remarkably, the interpretable DTs performed on par with, and sometimes better than, black-box algorithms, challenging the notion of a trade-off between performance and interpretability. These findings highlight the need for further research into interpretable solutions to meet industry demands. Unlike black-box approaches, which limit users to evaluate model outputs, interpretable models can enable interaction with their internal workings. For instance, integrating \textsc{eldt} with a user interface could allow domain experts to adjust DT values, fostering human-in-the-loop optimization. Additionally, incorporating Large Language Models into the proposed framework could further enhance interpretability providing natural language explanations of decision-making~\cite{serafim2024exploring}.


%
%
%
\newpage
\bibliographystyle{splncs} 
\bibliography{bibliography}


\end{document}